\documentclass[5p]{elsarticle} 

\usepackage{amssymb}
\usepackage{CJKutf8}
\usepackage{graphicx}
\usepackage{amsmath}
\usepackage{booktabs}
\usepackage{setspace}
\usepackage{multirow}
\usepackage{xcolor}
\usepackage[ruled,linesnumbered]{algorithm2e}

\usepackage{lineno}
\usepackage[colorlinks=true, allcolors=blue]{hyperref}

\journal{Biomedical Signal Processing and Control}

\begin{document}
\begin{CJK}{UTF8}{gbsn}

\begin{frontmatter}



\title{Self-training with dual uncertainty for semi-supervised medical image segmentation}


\author[address1]{Zhanhong Qiu}

\author[address1]{Haitao Gan \corref{cor1}}
\cortext[cor1]{Corresponding author}
\ead{htgan01@hbut.edu.cn}

\author[address1]{Ming Shi}

\author[address1]{Zhongwei Huang}

\author[address1]{Zhi Yang \corref{cor1}}
\ead{zyang631@hbut.edu.cn}

\address[address1]{School of Computer Science, Hubei University of Technology, Wuhan, Hubei 430068, China}

\begin{abstract}
In the field of semi-supervised medical image segmentation, the shortage of labeled data is the fundamental problem. How to effectively learn image features from unlabeled images to improve segmentation accuracy is the main research direction in this field. Traditional self-training methods can partially solve the problem of insufficient labeled data by generating pseudo labels for iterative training. However, noise generated due to the model's uncertainty during training directly affects the segmentation results. Therefore, we added sample-level and pixel-level uncertainty to stabilize the training process based on the self-training framework. Specifically, we saved several moments of the model during pre-training, and used the difference between their predictions on unlabeled samples as the sample-level uncertainty estimate for that sample. Then, we gradually add unlabeled samples from easy to hard during training. At the same time, we added a decoder with different upsampling methods to the segmentation network and used the difference between the outputs of the two decoders as pixel-level uncertainty. In short, we selectively retrained unlabeled samples and assigned pixel-level uncertainty to pseudo labels to optimize the self-training process. We compared the segmentation results of our model with five semi-supervised approaches on the public 2017 ACDC dataset and 2018 Prostate dataset. Our proposed method achieves better segmentation performance on both datasets under the same settings, demonstrating its effectiveness, robustness, and potential transferability to other medical image segmentation tasks.
\end{abstract}

\begin{keyword}
medical image segmentation \sep semi-supervised learning \sep self-training \sep uncertainty estimation
\end{keyword}

\end{frontmatter}

\section{Introduction}
\label{intro}

Image segmentation plays a critical role in medical image processing, facilitating diagnosis, lesion detection, surgical planning, and subsequent work. Traditional manual segmentation methods are time-consuming and rely heavily on the clinical experience of doctors, leading to inconsistent results. To enhance clinical efficiency and reliability, automatic segmentation techniques have been extensively studied. With the advent of deep learning, deep neural networks such as fully convolutional networks \cite{1} and U-net \cite{2} have achieved remarkable performance in automatic medical image segmentation. For instance, Attention U-net \cite{3} introduces an attention mechanism between the encoder and decoder to focus on regions of interest, while CE-net \cite{4} proposes an encoder-decoder that transforms image features into high-dimensional vectors for multi-scale feature learning. Tang et al. \cite{5} have also proposed an end-to-end medical image segmentation framework that incorporates uncertainty estimation to enhance model robustness. However, the high segmentation accuracy of these methods relies on a large amount of labeled data. Acquiring labeled data is expensive and time-consuming, and in practical applications, there is often a scarcity of labeled data. To tackle this problem, semi-supervised learning has been applied to medical image segmentation, utilizing a small amount of labeled data and a large amount of unlabeled data for learning. Currently, research in the field of semi-supervised medical image segmentation primarily focuses on utilizing unlabeled data to learn image features and improve segmentation accuracy.

\begin{figure}[!t]
    \centering
    \includegraphics[width=0.5\textwidth]{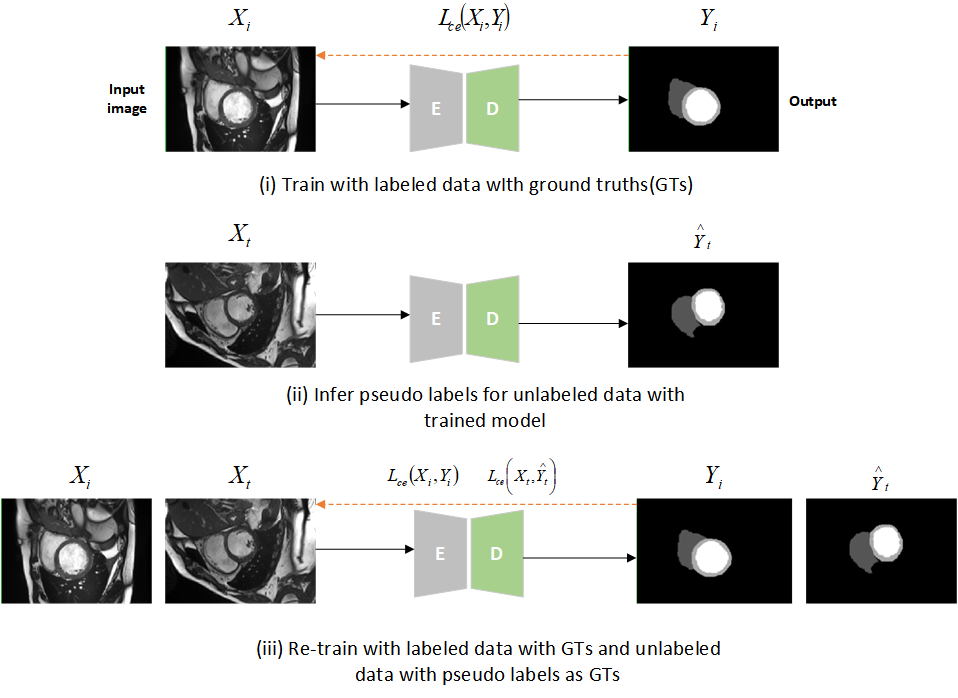}
    \caption{The overall process of self-training: (i) training an initial model using labeled data, (ii) inferring pseudo labels from the initial model, (iii) jointly training the model on labeled data and pseudo labeled data, and repeating steps (ii) and (iii) until convergence.}
    \label{fig:fig1}
\end{figure}

In the absence of sufficient labeled data, researchers have made multiple attempts to develop semi-supervised segmentation models. We have found that traditional semi-supervised segmentation models (see Figure\ref{fig:fig1}) mainly consist of the following stages: 1) Training an initial model using a small amount of labeled data, followed by predicting pseudo labels for the unlabeled data using the initial model \cite{6}; 2) Jointly iteratively training the final segmentation model using the labeled data with their ground truth, unlabeled data and their pseudo labels \cite{7,8}. However, the quality of the pseudo labels heavily impacts the accuracy of this approach, and low-quality predictions can amplify errors in subsequent training, leading to worse model performance. According to \cite{9}, pseudo labels with higher confidence are typically more effective in improving the segmentation accuracy.

Some methods employ SoftMax confidence \cite{10,11,12} to acquire high-quality pseudo labels. Specifically, they apply the SoftMax operation to the output of the network model, generating segmentation probability maps. Subsequently, a threshold is defined, classifying pixels with classification probabilities surpassing this threshold as foreground, while ignoring those below it. While this approach is straightforward and efficient, it is impractical to assign a uniform threshold to all pixels. At the same time, the Dropout method \cite{13} improves the accuracy of output results by randomly disabling some nodes during the training process. In addition, some methods leverage the reproducibility of the model \cite{14} by calculating the deviations between multiple predictions of the same image to obtain high-quality pseudo labels. Recently, some methods have achieved significant improvements by combining image features and predictions to calculate the uncertainty of pseudo labels \cite{15,16}.

Overall, the above methods still have limitations and shortcomings, such as the use of a single threshold, increased model complexity, and increased computational time. \textcolor{red}{Recently, Luo et al. \cite{17} used the distance between predictions at different scales in the U-Net model and the average prediction as an uncertainty estimation to rectify the consistency loss. Wu et al. \cite{18} extended the V-Net by adding two distinct decoders to obtain two sets of outputs. They then applied a sharpening function to the outputs to obtain soft pseudo labels. The consistency loss was computed based on the variance between the soft pseudo labels and the outputs to constrain the model. The methods of Luo et al.\cite{17} and Wu et al.\cite{18} is simple and effective. These two works use different predictions to impose consistency constraints on the model. However, these two methods primarily aim to reduce uncertainty in the model without applying it to predictions. In contrast, our method not only dynamically acquires uncertainty but also utilizes it to correct pseudo labels. Additionally, we have observed that the aforementioned methods do not take into account the differences between samples. In the case of medical images, there are variations in the size, shape, and location of lesions among different patients. To improve segmentation accuracy, it is beneficial to differentiate unlabeled samples to some extent.} Therefore, in this paper, we propose a new self-training-based network for semi-supervised medical image segmentation. First, we introduce sample-level uncertainty to differentiate between unlabeled samples based on their segmentation stability. And by conducting self-training on these unlabeled data in two stages based on priority, we make the entire training process smoother. At the same time, we add an additional decoder to the U-Net model to obtain two sets of different predictions. Then, we calculate the KL divergence between these two sets of predictions as an uncertainty estimation. Finally, we use this uncertainty to rectify the cross-entropy loss with pseudo labels as the target. Additionally, our model can adaptively update the pseudo labels and their uncertainties during the self-training process.

 In summary, our work makes three key contributions to the field of semi-supervised medical image segmentation.
1)	We proposed a new self-training framework for semi-supervised medical image segmentation based on uncertainty.
2)	We designed a method for sorting unlabeled samples by the stability of segmentation that can make the self-training process smoother.
3)	We designed a pixel-level uncertainty estimation method for rectifying cross-entropy loss.

Our model was evaluated on the Cardiac ACDC and Prostate datasets. In a semi-supervised setting, our model demonstrates superior performance compared to other methods, establishing a new state-of-the-art for self-training in semi-supervised medical image segmentation tasks.

\section{Related Works}
\label{rw}

\subsection{Semi-supervised learning}
With the advancement of deep learning, semi-supervised learning (SSL) has become a popular research direction in various computer vision tasks. Self-training methods \cite{19} are one of the most commonly used approaches, which iteratively trains network models using the predicted results (pseudo labels) of unlabeled data. The Π-model \cite{20} reduces noise in the learning process by enforcing consistency constraints between two different augmentations of the same sample. Similar to this, the mean-teacher \cite{21} model calculates the variance between the student-side outputs and the teacher-side outputs of the unlabeled images, adds the variance as a constraint to the loss function, and moves the weights of the student model towards the teacher model. The mean-teacher model is a simple and efficient method in SSL. At the same time, methods based on adversarial training \cite{22,23,24} enhance the robustness of models to unlabeled samples by discriminating between real and fake samples, thereby improving the performance of the models. Furthermore, data augmentation methods \cite{25,26} have been introduced to SSL, such as MixMatch \cite{27} and FixMatch \cite{28}, these methods effectively enhance the robustness of models by utilizing powerful data augmentation techniques to increase the diversity of data. In recent times, there has been a growing trend among semi-supervised learning methods to harness the synergistic interaction between two data representations and train them in a joint manner. \textcolor{red}{For example, the CoMatch \cite{29} framework introduces a novel architecture comprising a classification head and a projection head. The encoder is responsible for extracting image features, which are subsequently transformed into classification probabilities and embedding graph. Before being fed into the task heads, the images undergo both weak and strong augmentations to enhance their diversity. Notably, pseudo labels are constructed using the output of the weakly augmented images, providing supervision for the output of the strongly augmented images. By incorporating this approach, the model effectively learns image features across multiple dimensions, leading to outstanding performance. Similarly, SimMatch \cite{30} simultaneously maps image features to semantic space and instance space. Just like CoMatch \cite{29}, it leverages the output of weakly augmented images to generate pseudo labels. However, what distinguishes SimMatch \cite{30} is its utilization of the Unfold and Aggregate operations to seamlessly transform and merge image features across multiple spaces. This innovative approach has proven to be highly effective, leading to superior performance in comparison to CoMatch \cite{29}.} These prominent semi-supervised learning methods have been widely applied and improved in the field of medical imaging. We will discuss these in detail in Section 2.4. In this work, our main focus will be on the Self-training approach.

\subsection{Self-training}
Self-training is a method of training machine learning models using a small amount of labeled data and a large amount of unlabeled data. In this approach, initially, a pre-trained model is employed to make predictions on unlabeled data, generating pseudo labels as predicted outcomes for the unlabeled samples. Subsequently, these unlabeled data, along with their corresponding pseudo-labels, are incorporated into the training set and iteratively trained alongside the labeled data. During each training iteration, the model is updated and optimized using an augmented training set that encompasses both labeled and unlabeled data. This iterative process is repeated multiple times, progressively refining the model's performance and its ability to generalize to the unlabeled data. By training the model with a fusion of unlabeled data and their pseudo labels, self-training methods harness the information contained in the unlabeled data to improve the model's performance and extend its predictive and classifying capabilities to a broader range of samples. Pseudo-labeling was first introduced in 2013 \cite{6}, and since then, it has been applied and studied in various fields, such as image recognition \cite{31,32,33} and domain adaptation \cite{34,35}. In the semi-supervised setting, several methods have been proposed to improve self-training, such as using data augmentation and consistency loss functions \cite{36}, importance sampling based on confidence \cite{37}. In addition, strong data augmentation methods such as brightness, contrast, and colorization have been used to enhance the robustness of the model in semi-supervised semantic segmentation tasks \cite{38}.

These self-training methods typically use data augmentation and select high-confidence pseudo labels to improve segmentation accuracy, while we propose a method that sorts unlabeled samples based on the stability of segmentation and gradually adds them to the self-training process.

\subsection{Uncertainty estimation}
In deep learning, uncertainty can be divided into  aleatoric uncertainty and epistemic uncertainty. Aleatoric uncertainty is caused by annotation noise and is inherent in the data. On the other hand, epistemic uncertainty explains the differences between deep learning models and can be reduced through quantitative calculations. The research mentioned below regarding uncertainty computation refers to the estimation of epistemic uncertainty, which is the model uncertainty.

Until now, there have been several methods proposed to estimate model uncertainty. Bayesian networks \cite{39} have been widely used to estimate the uncertainty of the network. In Bayesian modeling, the MC-Dropout method \cite{40} was proposed, which uses dropout to sample multiple sub-models of the network. The differences between the outputs of different sub-models are used to represent the uncertainty of the model. Kendall et al. \cite{41} applied Bayesian theory to predict computer vision tasks and provided predictions of uncertainty. In addition, uncertainty has also been extensively studied in the field of medical image analysis \cite{41,42}. In particular, the method for estimating uncertainty mentioned in \cite{43} adds noise to the input to obtain two outputs. The mean of the two outputs is taken as the predicted result and the variance is used as the uncertainty map. Inspired by this work, we propose an adaptive method to automatically learn pixel-level uncertainty. Unlike the MC-Dropout method, we only perform one forward propagation to obtain the uncertainty map, which reduces computational costs and runtime.

\subsection{Semi-supervised medical image segmentation}
With the rapid advancement of semi-supervised learning, an increasing number of exceptional methods are continually emerging within the domain of medical image segmentation, consistently delivering outstanding results. Numerous semi-supervised learning methods discussed in Section 2.1 have found applications in the domain of semi-supervised medical image segmentation. In the following section, we will provide an introduction to a few of these methods.

Traditional self-training methods have garnered extensive utilization within the realm of semi-supervised medical image segmentation, demonstrating a multitude of notable application cases that highlight their effectiveness. For example, Bai et al. \cite{7} applied self-training methods in the semi-supervised segmentation of cardiac MRI images, and further refined the segmentation results by optimizing pseudo labels using conditional random fields. Wang et al. \cite{44} added a trust module to the self-training model to evaluate the pseudo labels and set a threshold to select high-confidence pixels. Li et al. \cite{45} proposed a self-ensembling strategy to construct accurate predictions using exponential moving average, thereby mitigating the impact of noisy and unstable pseudo labels. Thompson et al. \cite{46} introduced superpixel maps calculated by the simple linear iterative clustering (SLIC) algorithm to refine pseudo labels. This algorithm is particularly suitable for segmenting targets with irregular shapes. Shi et al. \cite{47} proposed a network called CoraNet, which obtains a conservative prediction and an aggressive prediction by setting the misclassification weight in the loss function, and the overlapping part of the two predictions is used as the pseudo label. The performance of self-training-based methods critically hinges on the quality of pseudo labels. Thus, acquiring high-quality pseudo labels is a pivotal aspect that demands careful consideration.

Adversarial learning-based methods foster the segmentation of unlabeled images towards the ground truth of labeled images. These techniques have also been successfully applied in the domain of semi-supervised medical image segmentation, showcasing their efficacy.  For instance, Zhang et al. \cite{48} incorporated adversarial learning to encourage the segmentation output of unlabeled data to closely resemble the annotations of labeled data. Chen et al. \cite{49} proposed the inclusion of a discriminator following the segmentation network. This discriminator is employed to differentiate between input signed distance maps originating from labeled images and unlabeled images. Wu et al. \cite{50} introduced the incorporation of two discriminators in their approach. One discriminator is utilized to predict confidence maps, while the other discriminator is employed to distinguish between the segmentation results obtained from labeled and unlabeled data. By incorporating an additional auxiliary discriminator, the issue of an undertrained primary discriminator caused by limited labeled images can be mitigated. However, achieving convergence in adversarial training can be a challenging task. 

Recently, consistency-based semi-supervised methods have received widespread attention due to their outstanding performance. The main idea of this approach is to encourage consistency in the outputs of different morphological variations of the same image. This method has found widespread application and improvement in the field of medical image segmentation. For example, Wang et al. \cite{51} extended the student-teacher model by incorporating segmentation uncertainty estimation and feature uncertainty estimation. Uncertainty was interactively used to modify the teacher model and also served as weights for the loss function to balance the training process. \textcolor{red}{Lin Hu et al. \cite{52} proposed a method for semi-supervised nasopharyngeal carcinoma (NPC) segmentation based on uncertainty and attention-guided consistency. The uncertainty is utilized to measure the reliability of predictions, enabling the model to emphasize the more reliable areas. A key contribution of their work is the introduction of a self-attention module, which guides the consistency loss by incorporating region of interest (ROI) images to capture more global information and constrain the model effectively.  Wang et al. \cite{53} expanded upon the mean teacher model by incorporating multi-task learning and contrastive learning techniques. They employed the mean teacher model alongside multi-task learning to promote both inter-task and intra-task consistency within the same input image, thus improving the model's robustness. Furthermore, they employed contrastive learning to differentiate between different views of samples, allowing the model to extract intrinsic features from diverse samples. Similarly to \cite{53}, Wang et al. \cite{54} also proposed a consistency-based approach among the tasks of segmentation, regression, and object detection to train their network model. Additionally, they incorporated a Boundary Box Attention (BBA) module, which focused the network's attention on the target regions. This module utilized boundary boxes to guide the model's attention and enhance its performance in localizing objects of interest.}

\section{Method}
\label{mt}

\begin{figure*}[!t]
    \centering
    \includegraphics[width=0.9\textwidth]{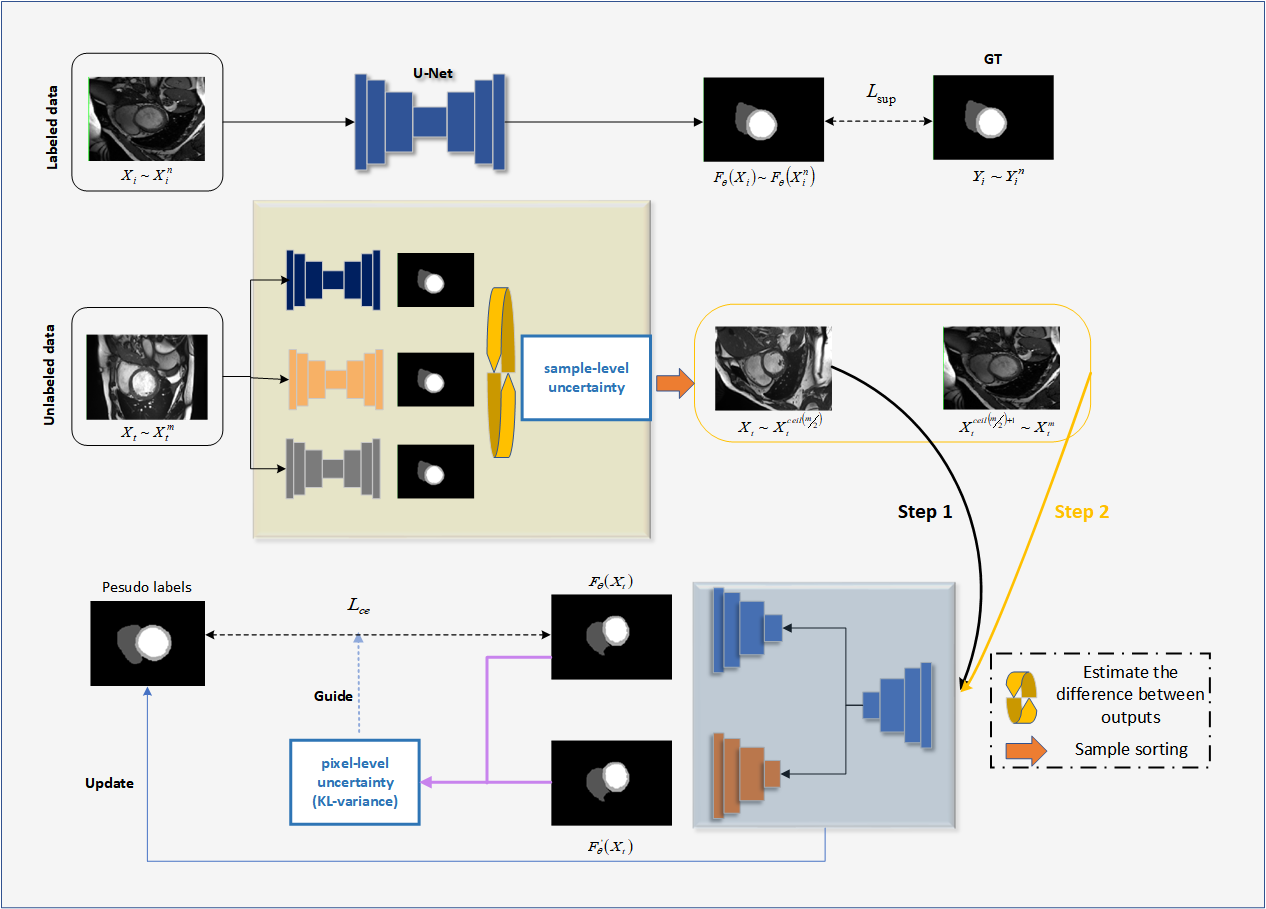}
    \caption{The main framework of our proposed method.}
    \label{fig:fig2}
\end{figure*}

In this section, we will introduce our method in detail. The proposed network is outlined in Figure \ref{fig:fig2}. In order to address the limitations of traditional pseudo label learning applied in the field of semi-supervised medical image segmentation. We introduce sample-level uncertainty and pixel-level uncertainty to stabilize the self-training process. First, we use K epochs of the model during the pre-training phase to estimate the uncertainty of the unlabeled samples. We then sort these unlabeled samples in ascending order based on their uncertainty. Next, we introduce the unlabeled samples into the self-training process in two stages to ensure a smoother training process. Secondly, we use the outputs of two different decoders to estimate the pixel-level uncertainty and use it to rectify the cross-entropy loss. Finally, we conduct iterative training to improve the performance of the model. The specific process of our proposed method is shown in Algorithm \ref{algorithm1}. To accurately describe this work, we provide the relevant setup of the problem: The entire dataset D is composed of the labeled dataset $D_l$ and the unlabeled dataset $D_u$, where the labeled dataset consists of pairs$(X_i,Y_i )$, where $Y_i$ is the ground truth for the labeled data $X_i$, and the unlabeled data is denoted by $X_t$. The network parameters are represented by $F(\theta)$.

\subsection{Sample sorting module with sample-level uncertainty}

In order to make the training process smoother, we divided the unlabeled samples into two subsets: reliable and unreliable, and gradually included them in the training process. Specifically, K models were saved at specific time points during the initial training process, denoted as $F_1 (\theta),F_2 (\theta),..., F_k (\theta)$  respectively. The unlabeled samples $X_t$ were then fed into these saved models to obtain the corresponding predictions $F_1 (X_t |\theta_t ),F_2 (X_t |\theta_t ),...,F_k (X_t |\theta_t )$. We estimate the sample-level uncertainty of each sample by using the differences between these predictions:

\begin{flalign}
\label{1}
\begin{split}
    & \text{Sample-level-uncertainty}= \\ & \sum_{i=1}^{\mathrm{k}-1} \operatorname{mean}\left[\left(F_i\left(X_t \mid \theta_t\right)-F_k\left(X_t \mid \theta_t\right)\right)^2\right] / \mathrm{k}-1 
\end{split}
\end{flalign}

We sorted the unlabeled samples by their sample-level uncertainty values in ascending order, and considered the first 50\% of samples as reliable and the remaining 50\% as relatively unreliable. These samples were added to the subsequent training in two separate stages. Specifically, in the first stage of self-training, labeled data and reliable unlabeled samples along with their pseudo labels are used for training. In the second stage of self-training, all data are included, and the model parameters from the first stage are inherited.

\subsection{Uncertainty rectified cross entropy loss}
In Section 2.3, we mentioned that the presence of model uncertainty introduces a certain degree of bias in predictions. In traditional self-training processes, the noise present in the obtained pseudo labels accumulates during iterations, thereby directly impacting the performance of the model. Meanwhile, this type of uncertainty can be reduced through estimation. \textcolor{red}{Inspired by previous work \cite{17,43}, We calculate the Kullback-Leibler divergence between the outputs of the two decoders in the model as an estimation of uncertainty. The Kullback-Leibler divergence (Relative Entropy) measures the distance between two random distributions. As the similarity between the two distributions increases, the entropy decreases, while the entropy increases as the gap between the two distributions widens. Specifically, in both Step 1 and Step 2 as illustrated in Figure 2, every unlabeled data undergoes two decoders to obtain the corresponding uncertainty.} Since we are calculating the distance between two images, we will obtain the entropy value for each pixel. To differentiate it from the sample-level uncertainty mentioned in Section 3.1, we refer to this type of uncertainty as pixel-level uncertainty. The specific pixel-level uncertainty (KL divergence) is shown in Eq. (\ref{2}).

\begin{flalign}
\label{2}
\begin{split}
    & \text{Pixel-level-uncertainty}= D_{kl} = \\ &F\left(X_t \mid \theta_t\right) \log \left[\frac{F\left(X_t \mid \theta_t\right)}{F_{aux}\left(X_t \mid \theta_t\right)}\right]
\end{split}
\end{flalign} '$F_{aux} (\theta)$'represents the parameters of the newly added decoder.

Traditional self-training methods employ cross-entropy loss to learn from unlabeled data \cite{7,45}. Specifically, the model's output is used as the prediction, and the pseudo labels are treated as the targets to minimize the loss (as shown in Eq. (\ref{3})).

\begin{equation}
\label{3}
L_{ce}=-\sum_{i=1}^n \hat{y}_t \log F\left(X_t \mid \theta_t\right)
\end{equation} Here, $\hat{y}_t$represents the pseudo label of unlabeled data and n is the number of segmentation classes in the dataset. Next, we use the uncertainty calculated by Eq. (\ref{2}) to rectify the cross-entropy loss with pseudo labels as the target. As mentioned above, $D_{kl}$ is a tensor that contains values for each pixel point. Similarly, the output of the cross-entropy function is also saved as the loss value for each pixel point. The rectified cross-entropy loss is shown in Eq. (\ref{4}).

\begin{equation}
\label{4}
L_{unsup}=\exp \left\{-D_{k l}\right\} L_{c e}+D_{k l}
\end{equation}

The unsupervised loss consists of two components: the rectified cross-entropy loss term and the uncertainty minimization term. The first term is based on $e^{-x}$, which assigns low weights to pixels with high uncertainty and high weights to pixels with low uncertainty. In addition, the second term is uncertainty estimation term that encourages two different decoders to generate more consistent predictions during the training process, aiming to reduce the model's uncertainty. the model will generate more accurate pseudo labels during the self-training process and adaptively acquire the corresponding uncertainty estimation term. 

\subsection{The overall loss function}
The proposed self-training framework (\ref{fig:fig2}) uses both the mean of the cross-entropy loss and the Dice Loss for supervised training in the pre-training stage:

\begin{equation}
\label{5}
L_{sup}=\frac{L_{dice}\left(X_i, Y_i\right)+L_{ce}\left(X_i, Y_i\right)}{2}
\end{equation} For re-training stage, both supervised and unsupervised losses are utilized to learn from labeled and unlabeled data:
\begin{equation}
\label{6}
L_{total}=L_{sup}+L_{unsup}
\end{equation}

\begin{algorithm}
\caption{The process of our proposed method}\label{algorithm1}
\begin{spacing}{1.25}
\KwData{Labeled training set $D_l=\left\{X_i, Y_i\right\}_{i=1}^N$; 
Unlabeled training set $D_u=\left\{X_t\right\}_{t=1}^M$;
Teacher/Student model $T/S$.}
\KwResult{Trained student model $S$.}
Train $T$ on $D_l$ and save $K$ models $\left\{F(\theta)_{i=1}^K\right\}$\;
Sort and partition the unlabeled samples based on Eq.(\ref{1})\;
$D_u=D_{u 1} \cup D_{u 2}$\;
Initialize $S$ with $T$ and generate pseudo labels for $D_{u1}$\;
Train $S$ on $D_l$ and $D_{u1}$ based on Eq.(\ref{6})\;
Update pseudo labels for $D_u$\;
Retrain $S$ on $D_l$ and $D_u$ based on Eq.(\ref{6})\;
\textbf{Return:}$S$\;
\end{spacing}
\end{algorithm}

\section{Experiments}
\label{eps}

\subsection{Datasets}
We evaluated the proposed method on two publicly available MRI datasets:\\
(a) ACDC dataset: It contains 100 short-axis MR-cine T1 3D volumes of cardiac anatomy acquired using 1.5T and 3T scanners. The expert annotations are provided for three structures: right ventricle, myocardium, and left ventricle. It was hosted as part of the MICCAI ACDC challenge 2017.\\
(b) Prostate dataset: This is a public dataset made available as part of the MICCAI 2018 medical segmentation decathlon challenge. It comprises 48 subjects T2 weighted MR scans of the prostate. The in-plane resolution ranges from 0.60x0.60mm to 0.75x0.75mm and through-plane resolution ranges from 2.99mm to 4mm. Segmentation masks comprise of two structures: peripheral zone (PZ) and central gland (CG). 

\begin{table*}[t] 
\centering 
\caption{Comparisons of the proposed method with other semi-supervised methods on the ACDC dataset. Reported values are averages for 5 runs with different random seeds.} 
\label{tab:tab1} 
\begin{tabular}{c|cc|cccc}\hline 
\multirow{2}{*}{\centering Method} &\multicolumn{2}{c|}{Scans used} &\multicolumn{4}{c}{Metrics}\\
\cline{2-7}
&Labeled &Unlabeled &Dice(\%) &Jaccard(\%) &95HD(voxel) &ASD(voxel) \\
\hline
U-net &7(10\%) &0 &77.34 &66.20 &12.18 &3.45 \\
U-net &14(20\%) &0 &81.35 &70.23 &10.74 &2.76 \\
U-net &All(70) &0 &91.65 &84.93 &1.89 &0.56 \\
\hline
MT(NIPS2017) \cite{21} &\multirow{6}{*}{7(10\%)}  &\multirow{6}{*}{63(90\%)} &81.53±0.30 &70.60±0.44 &10.04±1.54 &2.95±0.33\\
UA-MT(MICCAI2019) \cite{55} & & &81.80±0.34 &70.53±0.39 &10.61±1.38 &3.14±0.42\\
DCT-Seg (2020) \cite{56} & & &82.04±0.44 &71.22±0.46 &8.23±0.46 &2.61±0.15\\
URPC(MICCAI2021) \cite{17} & & &82.49±0.45 &71.96±0.18 &\textbf{6.06±0.81} &1.94±0.77\\
ST++(2022)  \cite{57} & & &80.14±0.36 &68.22±0.56 &14.05±2.69 &4.19±0.47\\
Ours & & &\textbf{83.44±0.60} &\textbf{72.60±0.86} &7.07±1.21 &\textbf{1.94±0.28}\\
\hline
MT(NIPS2017) \cite{21} &\multirow{6}{*}{14(20\%)}  &\multirow{6}{*}{56(80\%)} &84.61±0.74 &75.09±0.88 &9.38±2.81 &2.66±0.54\\
UA-MT(MICCAI2019) \cite{55} & & &84.76±0.91 &75.34±1.13 &7.42±1.32 &2.37±0.47\\
DCT-Seg (2020) \cite{56} & & &85.26±0.36 &75.88±0.62 &7.56±0.98 &2.29±0.17\\
URPC(MICCAI2021) \cite{17} & & &85.86±0.66 &76.65±0.90 &5.72±1.20 &1.68±0.34\\
ST++(2022)  \cite{57} & & &84.57±0.53 &74.45±0.91 &8.88±2.03 &2.47±0.52\\
Ours & & &\textbf{86.19±0.18} &\textbf{76.75±0.27} &\textbf{5.02±0.84} &\textbf{1.47±0.22}\\
\hline
\end{tabular}
\end{table*}

\begin{figure*}[h]
    \centering
    \includegraphics[width=1.0\textwidth]{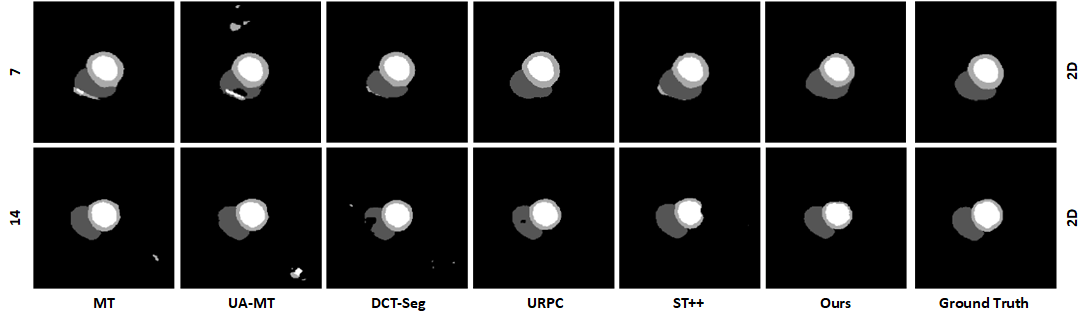}
    \caption{There are the segmentation results of MT[21], UA-MT[55], DCT-Seg[56], URPC[17], ST++[57], and our proposed method trained on 10\% and 20\% labeled data on the test images of the ACDC dataset, along with their corresponding ground truth. The black regions represent the background, while different shades of gray represent the three different foregrounds.}
    \label{fig:fig3}
\end{figure*}

\subsection{Implementing details}
Data split: For the ACDC dataset, we randomly selected 10 and 20 subjects as the validation and testing sets. respectively, while the remaining 70 subjects were used as the training set. For the Prostate dataset, we only evaluated the model using the 32 labeled subjects. We randomly divided the dataset into five parts, with three parts used as the training set, one part as the validation set, and the remaining part as the testing set. We conducted five-fold cross-validation. To enable proper semi-supervised training of the model, we divided the training set into labeled and unlabeled data, denoted as n/m, where n represents the number of labeled samples and m represents the number of unlabeled samples, with n significantly smaller than m.

Data preprocessing: We applied the same preprocessing to both datasets. Specifically, we normalized each 3D volume (x) during the generation of 2D slices. Afterwards, we enhanced the data by randomly rotating and flipping the 2D slices, and randomly cropping them to a size of 256×256.

Parameter settings: We used the SGD optimizer with a learning rate between 0.001 and 0.01, momentum of 0.9, and weight decay factor of 0.0001 for both datasets. During network training, the batch size was set to 8, and each training phase was iterated for 200 epochs. Our method was implemented using PyTorch on Nvidia GeForce RTX 3090 GPU.

Validation metrics: We used four metrics, including Dice, Jaccard, Average Surface Distance (ASD), and 95\% Hausdorff Distance (95HD), to quantitatively evaluate the performance of the model.

\begin{table*}[t] 
\centering  
\caption{Comparisons of the proposed method with other semi-supervised methods on the Prostate dataset. Reported values are averages for 5 runs with different data partitions.} 
\label{tab:tab2}
\begin{tabular}{c|cc|cccc}\hline 
\multirow{2}{*}{\centering Method} &\multicolumn{2}{c|}{Scans used} &\multicolumn{4}{c}{Metrics}\\
\cline{2-7}
&Labeled &Unlabeled &Dice(\%) &Jaccard(\%) &95HD(voxel) &ASD(voxel) \\
\hline
U-net &4 &0 &50.44 &37.71 &13.86 &4.59 \\
U-net &6 &0 &53.32 &39.87 &11.78 &3.80 \\
U-net &All(18) &0 &62.08 &48.77 &8.78 &2.63 \\
\hline
MT(NIPS2017) \cite{21} &\multirow{6}{*}{4}  &\multirow{6}{*}{14} &52.02±6.63 &38.16±5.47 &13.66±1.37 &4.41±0.64\\
UA-MT(MICCAI2019) \cite{55} & & &52.44±6.52 &38.39±5.53 &13.76±1.16 &4.16±0.38\\
DCT-Seg (2020) \cite{56} & & &52.92±6.32 &39.29±5.44 &12.51±1.49 &3.58±0.43\\
URPC(MICCAI2021) \cite{17} & & &53.36±6.38 &39.40±5.46 &11.96±1.77 &\textbf{3.41±0.54}\\
ST++(2022)  \cite{57} & & &52.63±6.29 &38.63±5.26 &13.15±2.06 &3.69±0.60\\
Ours & & &\textbf{54.13±6.54} &\textbf{40.41±5.72} &\textbf{11.92±1.63} &3.42±0.49\\
\hline
MT(NIPS2017) \cite{21} &\multirow{6}{*}{6}  &\multirow{6}{*}{12} &55.44±6.74 &42±5.65 &11.47±0.64 &3.17±0.41\\
UA-MT(MICCAI2019) \cite{55} & & &55.59±6.83 &42.27±5.76 &11.69±1.55 &3.43±0.68\\
DCT-Seg (2020) \cite{56} & & &56.55±7.16 &43.18±6.11 &10.24±1.31 &2.99±0.36\\
URPC(MICCAI2021) \cite{17} & & &56.61±6.92 &43.22±6.03 &9.80±1.08 &2.87±0.37\\
ST++(2022)  \cite{57} & & &55.94±6.60 &42.43±5.71 &11.58±0.53 &3.27±0.34\\
Ours & & &\textbf{57.78±6.38} &\textbf{44.10±5.85} &\textbf{9.65±0.61} &\textbf{2.77±0.37}\\
\hline
\end{tabular}
\end{table*}

\begin{figure*}[h]
    \centering
    \includegraphics[width=1.0\textwidth]{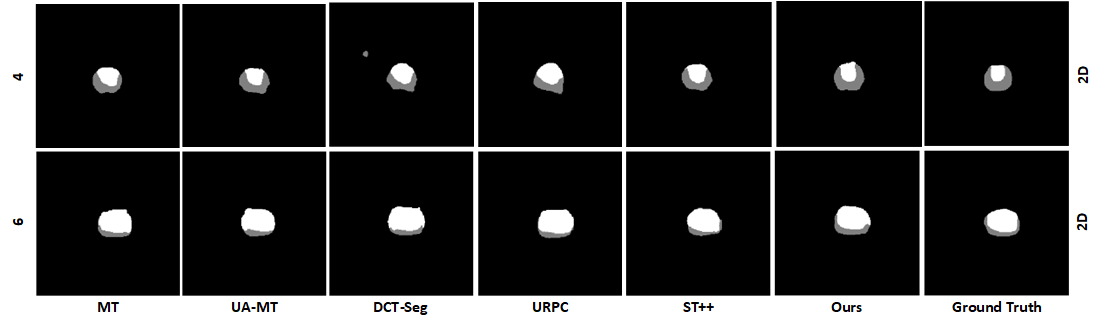}
    \caption{There are the segmentation results of MT[21], UA-MT[55], DCT-Seg[56], URPC[17], ST++[57], and our proposed method trained on 4 and 6 labeled data on the test images of the Prostate dataset, along with their corresponding ground truth. The black regions represent the background, while different shades of gray represent the two different foregrounds.}
    \label{fig:fig4}
\end{figure*}

\section{Result}
\label{rst}

\subsection{Results on the ACDC dataset}
Table \ref{tab:tab1} presents the performance of our model and five semi-supervised methods on the ACDC dataset, with the average of three segmentation targets (myocardium, left ventricle, and right ventricle) used as the final result. As observed, our model achieved the highest Dice and Jaccard scores at the 10\% and 20\% labeled data settings compared to other methods. We also achieved excellent performance in terms of the ASD and 95HD metrics. Specifically, compared to UA-MT\cite{55} and URPC\cite{17} models that have uncertainty estimation, our method still achieves better results in both semi-supervised settings. Meanwhile, compared to fully supervised U-Net models with 10\% and 20\% labeled data, our method can leverage the unlabeled data and largely improve Dice from 77.34\%, 81.35\% to 83.44\%, 86.19\%, respectively. It can be observed that as the amount of labeled data increases, the relative improvement of the semi-supervised method compared to the fully supervised approach gradually decreases.

Additionally, Figure \ref{fig:fig3} provides several segmentation results of two images on the ACDC dataset. They were obtained from five comparative models and our proposed method. It can be observed that our method accurately segments the contour of the target region, rather than considering areas outside the contour as foreground (e.g., MT\cite{21}, UAMT\cite{55}, DCT-Seg\cite{56}). Moreover, our segmentation results exhibit smoother edges and closely resemble the ground truth. Additionally, in the case of only 10\% labeled data, the comparative results show instances where the foreground is incorrectly segmented into wrong categories. However, our segmentation results do not exhibit such cases.

\subsection{Results on the  Prostate dataset}

To verify the applicability of the proposed method on other datasets, we further conducted the same comparison experiments on the Prostate dataset. As mentioned in Section 4.1, the Prostate dataset contains valid data from only 32 patients. To validate the effectiveness of the model, we performed five-fold cross-validation, and the final statistical results and visualization are presented in Table \ref{tab:tab2} and Figure \ref{fig:fig4}. Similarly, in Table \ref{tab:tab2}, the proposed method achieved the highest Dice and Jaccard scores in both semi-supervised settings, while the performance of URPC\cite{17} was close to ours in terms of 95HD and ASD metrics. Compared to the fully supervised baseline, our method improved Dice and Jaccard scores from 53.32\% and 39.87\% to 57.78\% and 44.10\%, respectively, when n=6. This direct performance improvement is attributed to semi-supervised learning.

From Figure \ref{fig:fig4}, it can be observed that the segmentation target regions in the Prostate dataset are relatively small, which leads to lower overall segmentation accuracy and more noise in the predicted results. However, compared to other methods, our approach still achieves predictions that are closest to the ground truth. Specifically, our method produces coherent and continuous predictions, resulting in smoother edges and a better adherence to the organ structures.

\textcolor{red}{To validate the statistical significance of the improvements achieved by our method compared to others, we conducted T-tests on all comparative results from the two datasets. The specific results are presented in Table \ref{tab:tab3} and Table \ref{tab:tab4}, The results indicate that in the majority of settings, our method exhibits significant advantages over the comparative methods (P\textless0.05). As the amount of labeled data increases, the differences between our method and some individual methods become less pronounced. Overall, our method demonstrates a certain degree of superiority compared to the previous approaches.}

In short, based on the results from two datasets, the proposed method has been demonstrated to be effective and applicable, showcasing its validity and generalizability. In particular, we achieved segmentation performance comparable to that of unsupervised regularization-based methods using the traditional self-training approach. Additionally, our model does not rely on specific backbones and can be applied for many medical image segmentation tasks. Most importantly, our method effectively leverages unlabeled data to enhance segmentation performance.

\begin{table}[h] 
\centering 
\caption{The paired T-test results (P-values) of Dice scores obtained from the comparative experiments in the two semi-supervised settings on the ACDC dataset.} 
\label{tab:tab3} 
\begin{small}
\begin{tabular}{c|c|c|c|c|c} 
\hline 
  &MT &UA-MT &DCT-Seg &URPC &ST++ \\ 
\hline
7 labeled  &0.0028 &0.0014 &0.0144 &0.0285 &0.0040\\
\hline
14 labeled  &0.0157 &0.0530 &0.0098 &0.1660 &0.0122\\
\hline
\end{tabular}
\end{small}
\end{table}

\begin{table}[h] 
\centering 
\caption{The paired T-test results (P-values) of Dice scores obtained from the comparative experiments in the two semi-supervised settings on the Prostate dataset.} 
\label{tab:tab4} 
\begin{small}
\begin{tabular}{c|c|c|c|c|c} 
\hline 
   &MT &UA-MT &DCT-Seg &URPC &ST++ \\ 
\hline
4 labeled  &0.0001 &0.0023 &0.0331 &0.0054 &0.0031\\
\hline
6 labeled  &0.0013 &0.0012 &0.0347 &0.089 &0.0005\\
\hline
\end{tabular}
\end{small}
\end{table}

\subsection{Ablation experiments}
In the introduction, we mentioned that our model integrates sample-level uncertainty and pixel-level uncertainty into the self-training process to enhance training. To validate the contributions of these two modules, we conducted a series of experiments on various model settings using the 2017 ACDC dataset.

Our experiments compared (1) the traditional self-training method, (2) the self-training method with sample-level uncertainty, and (3) the self-training method with both sample-level and pixel-level uncertainty. The quantitative results are presented in Table \ref{tab:tab5}. As shown in the results, the traditional self-training methods update pseudo-labels during the iterative process, enabling them to make use of unlabeled data to some extent. Specifically, in both semi-supervised settings, the traditional self-training method improved Dice by approximately 2\% compared to the fully supervised baseline.

By incorporating sample-level uncertainty, we observed varying degrees of improvement in the Dice and Jaccard indices, indicating that our method of distinguishing between unlabeled samples positively impacted segmentation performance. After incorporating pixel-level uncertainty, the predictions showed significant improvements across all four metrics. It is worth noting that our predictions exhibited a decrease of 4 and 1 in the 95HD and ASD metrics, respectively (where 95HD and ASD represent the boundary distance between predicted and ground truth images, and lower values are preferable). This suggests that the proposed pixel-level uncertainty effectively estimates pixels located at the boundaries in the images. Subsequently, these pixels are correctly classified into the appropriate categories through the uncertainty rectified cross entropy loss module. Additionally, the visualization results in Figures \ref{fig:fig3} and Figures \ref{fig:fig4} also confirm this finding.

Through ablation experiments, we validated the effectiveness of the proposed modules and demonstrated significant improvements in our final model over the traditional self-training method.

\begin{table*}[h] 
\centering 
\caption{Ablation analysis on key modules on 2017 ACDC dataset. Reported values are averages for 5 runs with different random seeds.} 
\label{tab:tab5} 
\begin{tabular}{c|cc|cccc}\hline 
\multirow{2}{*}{\centering Method} &\multicolumn{2}{c|}{Scans used} &\multicolumn{4}{c}{Metrics}\\
\cline{2-7}
&Labeled &Unlabeled &Dice(\%) &Jaccard(\%) &95HD(voxel) &ASD(voxel) \\
\hline
U-net &7(10\%) &0 &77.34 &66.20 &12.18 &3.45 \\
U-net &14(20\%) &0 &81.35 &70.23 &10.74 &2.76 \\
U-net &All(70) &0 &91.65 &84.93 &1.89 &0.56 \\
\hline
Self-Training  &\multirow{3}{*}{7(10\%)}  &\multirow{3}{*}{63(90\%)} &79.45±0.58 &67.51±0.82 &11.36±1.53 &3.10±0.37\\
ST+sample-level & & &81.05±0.94 &69.50±1.12 &11.07±2.35 &3.42±0.62\\
ST+sample-pixel-level & & &\textbf{83.44±0.60} &\textbf{72.60±0.86} &7.07±1.21 &\textbf{1.94±0.28}\\
\hline
Self-Training  &\multirow{3}{*}{14(20\%)}  &\multirow{3}{*}{56(80\%)} &83.50±1.01 &73.13±1.30 &10.50±3.96 &3.10±0.95\\
ST+sample-level & & &84.90±0.49 &74.85±0.74 &9.94±1.30 &2.95±0.35\\
ST+sample-pixel-level & & &\textbf{86.19±0.18} &\textbf{76.75±0.27} &\textbf{5.02±0.84} &\textbf{1.47±0.22}\\
\hline
\end{tabular}
\end{table*}

\section{Conclusion}

In this paper, to address the limitation of insufficient labeled data, we proposed a self-training method with dual uncertainty for semi-supervised medical image segmentation. The method effectively resolves the issue of excessive noise generation during the iteration process in traditional self-training methods, thereby better utilizing unlabeled data to improve segmentation performance.  Specifically, based on traditional self-training methods, we proposed sample-level uncertainty to measure the segmentation difficulty of each sample. Subsequently, we divided the unlabeled data into two groups based on their segmentation difficulty and incorporated them into the training process in a smoother manner. Additionally, to obtain high-quality pseudo labels, we proposed pixel-level uncertainty to estimate the classification probability for each pixel. We assigned corresponding weights to each pixel to reduce noise during the learning process. The experimental results on two public medical datasets demonstrated the effectiveness and generalizability of this method.

\section*{Acknowledge}

This work is supported by the High-level Talents Fund of Hubei University of Technology under grant No.GCRC2020016, Doctoral Scientific Research Foundation of Hubei University of Technology under grant No.BSDQ2020064, National Natural Science Foundation of China under grant No. 62201203 and 61601162, Natural Science Foundation of Hubei Province under grant No.2021CFB282.


%
%
%

\end{CJK}

\begin{thebibliography}{10}

\bibitem{1}
Jonathan Long, Evan Shelhamer, and Trevor Darrell.
\newblock Fully convolutional networks for semantic segmentation.
\newblock In {\em Proceedings of the IEEE conference on computer vision and
  pattern recognition}, pages 3431--3440, 2015.

\bibitem{2}
Olaf Ronneberger, Philipp Fischer, and Thomas Brox.
\newblock U-net: Convolutional networks for biomedical image segmentation.
\newblock In {\em Medical Image Computing and Computer-Assisted
  Intervention--MICCAI 2015: 18th International Conference, Munich, Germany,
  October 5-9, 2015, Proceedings, Part III 18}, pages 234--241. Springer, 2015.

\bibitem{3}
Ozan Oktay, Jo~Schlemper, Loic~Le Folgoc, Matthew Lee, Mattias Heinrich,
  Kazunari Misawa, Kensaku Mori, Steven McDonagh, Nils~Y Hammerla, Bernhard
  Kainz, et~al.
\newblock Attention u-net: Learning where to look for the pancreas.
\newblock {\em arXiv preprint arXiv:1804.03999}, 2018.

\bibitem{4}
Zaiwang Gu, Jun Cheng, Huazhu Fu, Kang Zhou, Huaying Hao, Yitian Zhao, Tianyang
  Zhang, Shenghua Gao, and Jiang Liu.
\newblock Ce-net: Context encoder network for 2d medical image segmentation.
\newblock {\em IEEE transactions on medical imaging}, 38(10):2281--2292, 2019.

\bibitem{5}
Pin Tang, Pinli Yang, Dong Nie, Xi~Wu, Jiliu Zhou, and Yan Wang.
\newblock Unified medical image segmentation by learning from uncertainty in an
  end-to-end manner.
\newblock {\em Knowledge-Based Systems}, 241:108215, 2022.

\bibitem{6}
Dong-Hyun Lee et~al.
\newblock Pseudo-label: The simple and efficient semi-supervised learning
  method for deep neural networks.
\newblock In {\em Workshop on challenges in representation learning, ICML},
  volume~3, page 896. Atlanta, 2013.

\bibitem{7}
Wenjia Bai, Ozan Oktay, Matthew Sinclair, Hideaki Suzuki, Martin Rajchl,
  Giacomo Tarroni, Ben Glocker, Andrew King, Paul~M Matthews, and Daniel
  Rueckert.
\newblock Semi-supervised learning for network-based cardiac mr image
  segmentation.
\newblock In {\em Medical Image Computing and Computer-Assisted Intervention-
  MICCAI 2017: 20th International Conference, Quebec City, QC, Canada,
  September 11-13, 2017, Proceedings, Part II 20}, pages 253--260. Springer,
  2017.

\bibitem{8}
Deng-Ping Fan, Tao Zhou, Ge-Peng Ji, Yi~Zhou, Geng Chen, Huazhu Fu, Jianbing
  Shen, and Ling Shao.
\newblock Inf-net: Automatic covid-19 lung infection segmentation from ct
  images.
\newblock {\em IEEE transactions on medical imaging}, 39(8):2626--2637, 2020.

\bibitem{9}
Xiaoyan Wang, Yiwen Yuan, Dongyan Guo, Xiaojie Huang, Ying Cui, Ming Xia,
  Zhenhua Wang, Cong Bai, and Shengyong Chen.
\newblock Ssa-net: Spatial self-attention network for covid-19 pneumonia
  infection segmentation with semi-supervised few-shot learning.
\newblock {\em Medical Image Analysis}, 79:102459, 2022.

\bibitem{10}
Shuai Chen, Gerda Bortsova, Antonio Garc{\'\i}a-Uceda~Ju{\'a}rez, Gijs
  Van~Tulder, and Marleen De~Bruijne.
\newblock Multi-task attention-based semi-supervised learning for medical image
  segmentation.
\newblock In {\em Medical Image Computing and Computer Assisted
  Intervention--MICCAI 2019: 22nd International Conference, Shenzhen, China,
  October 13--17, 2019, Proceedings, Part III 22}, pages 457--465. Springer,
  2019.

\bibitem{11}
Liyan Sun, Jianxiong Wu, Xinghao Ding, Yue Huang, Guisheng Wang, and Yizhou Yu.
\newblock A teacher-student framework for semi-supervised medical image
  segmentation from mixed supervision.
\newblock {\em arXiv preprint arXiv:2010.12219}, 2020.

\bibitem{12}
Kevinminh Ta, Shawn~S Ahn, John~C Stendahl, Albert~J Sinusas, and James~S
  Duncan.
\newblock A semi-supervised joint network for simultaneous left ventricular
  motion tracking and segmentation in 4d echocardiography.
\newblock In {\em Medical Image Computing and Computer Assisted
  Intervention--MICCAI 2020: 23rd International Conference, Lima, Peru, October
  4--8, 2020, Proceedings, Part VI 23}, pages 468--477. Springer, 2020.

\bibitem{13}
Nitish Srivastava, Geoffrey Hinton, Alex Krizhevsky, Ilya Sutskever, and Ruslan
  Salakhutdinov.
\newblock Dropout: a simple way to prevent neural networks from overfitting.
\newblock {\em The journal of machine learning research}, 15(1):1929--1958,
  2014.

\bibitem{14}
Tanya Nair, Doina Precup, Douglas~L Arnold, and Tal Arbel.
\newblock Exploring uncertainty measures in deep networks for multiple
  sclerosis lesion detection and segmentation.
\newblock {\em Medical image analysis}, 59:101557, 2020.

\bibitem{15}
Xinyu Zhang, Jiewei Cao, Chunhua Shen, and Mingyu You.
\newblock Self-training with progressive augmentation for unsupervised
  cross-domain person re-identification.
\newblock In {\em Proceedings of the IEEE/CVF International Conference on
  Computer Vision}, pages 8222--8231, 2019.

\bibitem{16}
Eric Arazo, Diego Ortego, Paul Albert, Noel~E O’Connor, and Kevin McGuinness.
\newblock Pseudo-labeling and confirmation bias in deep semi-supervised
  learning.
\newblock In {\em 2020 International Joint Conference on Neural Networks
  (IJCNN)}, pages 1--8. IEEE, 2020.

\bibitem{17}
Xiangde Luo, Wenjun Liao, Jieneng Chen, Tao Song, Yinan Chen, Shichuan Zhang,
  Nianyong Chen, Guotai Wang, and Shaoting Zhang.
\newblock Efficient semi-supervised gross target volume of nasopharyngeal
  carcinoma segmentation via uncertainty rectified pyramid consistency.
\newblock In {\em Medical Image Computing and Computer Assisted
  Intervention--MICCAI 2021: 24th International Conference, Strasbourg, France,
  September 27--October 1, 2021, Proceedings, Part II 24}, pages 318--329.
  Springer, 2021.

\bibitem{18}
Yicheng Wu, Minfeng Xu, Zongyuan Ge, Jianfei Cai, and Lei Zhang.
\newblock Semi-supervised left atrium segmentation with mutual consistency
  training.
\newblock In {\em Medical Image Computing and Computer Assisted
  Intervention--MICCAI 2021: 24th International Conference, Strasbourg, France,
  September 27--October 1, 2021, Proceedings, Part II 24}, pages 297--306.
  Springer, 2021.

\bibitem{19}
Shengxing Liu and Fei-Yun Wu.
\newblock Self-training dictionary based approximated l0 norm constraint
  reconstruction for compressed ecg.
\newblock {\em Biomedical Signal Processing and Control}, 68:102768, 2021.

\bibitem{20}
Samuli Laine and Timo Aila.
\newblock Temporal ensembling for semi-supervised learning.
\newblock {\em arXiv preprint arXiv:1610.02242}, 2016.

\bibitem{21}
Antti Tarvainen and Harri Valpola.
\newblock Mean teachers are better role models: Weight-averaged consistency
  targets improve semi-supervised deep learning results.
\newblock {\em Advances in neural information processing systems}, 30, 2017.

\bibitem{22}
Takeru Miyato, Shin-ichi Maeda, Masanori Koyama, and Shin Ishii.
\newblock Virtual adversarial training: a regularization method for supervised
  and semi-supervised learning.
\newblock {\em IEEE transactions on pattern analysis and machine intelligence},
  41(8):1979--1993, 2018.

\bibitem{23}
Sudhanshu Mittal, Maxim Tatarchenko, and Thomas Brox.
\newblock Semi-supervised semantic segmentation with high-and low-level
  consistency.
\newblock {\em IEEE transactions on pattern analysis and machine intelligence},
  43(4):1369--1379, 2019.

\bibitem{24}
You Chenyu, Ruihan Zhao, Fenglin Liu, Sandeep Chinchali, Ufuk Topcu, Lawrence
  Staib, and James~S Duncan.
\newblock Class-aware generative adversarial transformers for medical image
  segmentation.
\newblock {\em arXiv preprint arXiv:2201.10737}, 2022.

\bibitem{25}
Ekin~D Cubuk, Barret Zoph, Dandelion Mane, Vijay Vasudevan, and Quoc~V Le.
\newblock Autoaugment: Learning augmentation strategies from data.
\newblock In {\em Proceedings of the IEEE/CVF conference on computer vision and
  pattern recognition}, pages 113--123, 2019.

\bibitem{26}
Ekin~D Cubuk, Barret Zoph, Jonathon Shlens, and Quoc~V Le.
\newblock Randaugment: Practical automated data augmentation with a reduced
  search space.
\newblock In {\em Proceedings of the IEEE/CVF conference on computer vision and
  pattern recognition workshops}, pages 702--703, 2020.

\bibitem{27}
David Berthelot, Nicholas Carlini, Ian Goodfellow, Nicolas Papernot, Avital
  Oliver, and Colin~A Raffel.
\newblock Mixmatch: A holistic approach to semi-supervised learning.
\newblock {\em Advances in neural information processing systems}, 32, 2019.

\bibitem{28}
Kihyuk Sohn, David Berthelot, Nicholas Carlini, Zizhao Zhang, Han Zhang,
  Colin~A Raffel, Ekin~Dogus Cubuk, Alexey Kurakin, and Chun-Liang Li.
\newblock Fixmatch: Simplifying semi-supervised learning with consistency and
  confidence.
\newblock {\em Advances in neural information processing systems}, 33:596--608,
  2020.

\bibitem{29}
Junnan Li, Caiming Xiong, and Steven~CH Hoi.
\newblock Comatch: Semi-supervised learning with contrastive graph
  regularization.
\newblock In {\em Proceedings of the IEEE/CVF International Conference on
  Computer Vision}, pages 9475--9484, 2021.

\bibitem{30}
Mingkai Zheng, Shan You, Lang Huang, Fei Wang, Chen Qian, and Chang Xu.
\newblock Simmatch: Semi-supervised learning with similarity matching.
\newblock In {\em Proceedings of the IEEE/CVF Conference on Computer Vision and
  Pattern Recognition}, pages 14471--14481, 2022.

\bibitem{31}
Qizhe Xie, Minh-Thang Luong, Eduard Hovy, and Quoc~V Le.
\newblock Self-training with noisy student improves imagenet classification.
\newblock In {\em Proceedings of the IEEE/CVF conference on computer vision and
  pattern recognition}, pages 10687--10698, 2020.

\bibitem{32}
I~Zeki Yalniz, Herv{\'e} J{\'e}gou, Kan Chen, Manohar Paluri, and Dhruv
  Mahajan.
\newblock Billion-scale semi-supervised learning for image classification.
\newblock {\em arXiv preprint arXiv:1905.00546}, 2019.

\bibitem{33}
Barret Zoph, Golnaz Ghiasi, Tsung-Yi Lin, Yin Cui, Hanxiao Liu, Ekin~Dogus
  Cubuk, and Quoc Le.
\newblock Rethinking pre-training and self-training.
\newblock {\em Advances in neural information processing systems},
  33:3833--3845, 2020.

\bibitem{34}
Yang Zou, Zhiding Yu, BVK Kumar, and Jinsong Wang.
\newblock Unsupervised domain adaptation for semantic segmentation via
  class-balanced self-training.
\newblock In {\em Proceedings of the European conference on computer vision
  (ECCV)}, pages 289--305, 2018.

\bibitem{35}
Ananya Kumar, Tengyu Ma, and Percy Liang.
\newblock Understanding self-training for gradual domain adaptation.
\newblock In {\em International Conference on Machine Learning}, pages
  5468--5479. PMLR, 2020.

\bibitem{36}
Qizhe Xie, Zihang Dai, Eduard Hovy, Thang Luong, and Quoc Le.
\newblock Unsupervised data augmentation for consistency training.
\newblock {\em Advances in neural information processing systems},
  33:6256--6268, 2020.

\bibitem{37}
Eric Arazo, Diego Ortego, Paul Albert, Noel~E O’Connor, and Kevin McGuinness.
\newblock Pseudo-labeling and confirmation bias in deep semi-supervised
  learning.
\newblock In {\em 2020 International Joint Conference on Neural Networks
  (IJCNN)}, pages 1--8. IEEE, 2020.

\bibitem{38}
Jianlong Yuan, Yifan Liu, Chunhua Shen, Zhibin Wang, and Hao Li.
\newblock A simple baseline for semi-supervised semantic segmentation with
  strong data augmentation.
\newblock In {\em Proceedings of the IEEE/CVF International Conference on
  Computer Vision}, pages 8229--8238, 2021.

\bibitem{39}
Finn~V Jensen and Thomas~Dyhre Nielsen.
\newblock {\em Bayesian networks and decision graphs}, volume~2.
\newblock Springer, 2007.

\bibitem{40}
Yarin Gal and Zoubin Ghahramani.
\newblock Dropout as a bayesian approximation: Representing model uncertainty
  in deep learning.
\newblock In {\em international conference on machine learning}, pages
  1050--1059. PMLR, 2016.

\bibitem{41}
Alex Kendall and Yarin Gal.
\newblock What uncertainties do we need in bayesian deep learning for computer
  vision?
\newblock {\em Advances in neural information processing systems}, 30, 2017.

\bibitem{42}
Zach Eaton-Rosen, Felix Bragman, Sotirios Bisdas, S{\'e}bastien Ourselin, and
  M~Jorge Cardoso.
\newblock Towards safe deep learning: accurately quantifying biomarker
  uncertainty in neural network predictions.
\newblock In {\em Medical Image Computing and Computer Assisted
  Intervention--MICCAI 2018: 21st International Conference, Granada, Spain,
  September 16-20, 2018, Proceedings, Part I}, pages 691--699. Springer, 2018.

\bibitem{43}
Alain Jungo and Mauricio Reyes.
\newblock Assessing reliability and challenges of uncertainty estimations for
  medical image segmentation.
\newblock In {\em Medical Image Computing and Computer Assisted
  Intervention--MICCAI 2019: 22nd International Conference, Shenzhen, China,
  October 13--17, 2019, Proceedings, Part II 22}, pages 48--56. Springer, 2019.

\bibitem{44}
Xiaoyan Wang, Yiwen Yuan, Dongyan Guo, Xiaojie Huang, Ying Cui, Ming Xia,
  Zhenhua Wang, Cong Bai, and Shengyong Chen.
\newblock Ssa-net: Spatial self-attention network for covid-19 pneumonia
  infection segmentation with semi-supervised few-shot learning.
\newblock {\em Medical Image Analysis}, 79:102459, 2022.

\bibitem{45}
Caizi Li, Li~Dong, Qi~Dou, Fan Lin, Kebao Zhang, Zuxin Feng, Weixin Si, Xuesong
  Deng, Zhe Deng, and Pheng-Ann Heng.
\newblock Self-ensembling co-training framework for semi-supervised covid-19 ct
  segmentation.
\newblock {\em IEEE Journal of Biomedical and Health Informatics},
  25(11):4140--4151, 2021.

\bibitem{46}
Bethany~H Thompson, Gaetano Di~Caterina, and Jeremy~P Voisey.
\newblock Pseudo-label refinement using superpixels for semi-supervised brain
  tumour segmentation.
\newblock In {\em 2022 IEEE 19th International Symposium on Biomedical Imaging
  (ISBI)}, pages 1--5. IEEE, 2022.

\bibitem{47}
Yinghuan Shi, Jian Zhang, Tong Ling, Jiwen Lu, Yefeng Zheng, Qian Yu, Lei Qi,
  and Yang Gao.
\newblock Inconsistency-aware uncertainty estimation for semi-supervised
  medical image segmentation.
\newblock {\em IEEE transactions on medical imaging}, 41(3):608--620, 2021.

\bibitem{48}
Yizhe Zhang, Lin Yang, Jianxu Chen, Maridel Fredericksen, David~P Hughes, and
  Danny~Z Chen.
\newblock Deep adversarial networks for biomedical image segmentation utilizing
  unannotated images.
\newblock In {\em Medical Image Computing and Computer Assisted Intervention-
  MICCAI 2017: 20th International Conference, Quebec City, QC, Canada,
  September 11-13, 2017, Proceedings, Part III 20}, pages 408--416. Springer,
  2017.

\bibitem{49}
Gaoxiang Chen, Jintao Ru, Yilin Zhou, Islem Rekik, Zhifang Pan, Xiaoming Liu,
  Yezhi Lin, Beichen Lu, and Jialin Shi.
\newblock Mtans: Multi-scale mean teacher combined adversarial network with
  shape-aware embedding for semi-supervised brain lesion segmentation.
\newblock {\em NeuroImage}, 244:118568, 2021.

\bibitem{50}
Huisi Wu, Guilian Chen, Zhenkun Wen, and Jing Qin.
\newblock Collaborative and adversarial learning of focused and dispersive
  representations for semi-supervised polyp segmentation.
\newblock In {\em Proceedings of the IEEE/CVF International Conference on
  Computer Vision}, pages 3489--3498, 2021.

\bibitem{51}
Yixin Wang, Yao Zhang, Jiang Tian, Cheng Zhong, Zhongchao Shi, Yang Zhang, and
  Zhiqiang He.
\newblock Double-uncertainty weighted method for semi-supervised learning.
\newblock In {\em Medical Image Computing and Computer Assisted
  Intervention--MICCAI 2020: 23rd International Conference, Lima, Peru, October
  4--8, 2020, Proceedings, Part I 23}, pages 542--551. Springer, 2020.

\bibitem{52}
Lin Hu, Jiaxin Li, Xingchen Peng, Jianghong Xiao, Bo~Zhan, Chen Zu, Xi~Wu,
  Jiliu Zhou, and Yan Wang.
\newblock Semi-supervised npc segmentation with uncertainty and attention
  guided consistency.
\newblock {\em Knowledge-Based Systems}, 239:108021, 2022.

\bibitem{53}
Kaiping Wang, Bo~Zhan, Chen Zu, Xi~Wu, Jiliu Zhou, Luping Zhou, and Yan Wang.
\newblock Semi-supervised medical image segmentation via a tripled-uncertainty
  guided mean teacher model with contrastive learning.
\newblock {\em Medical Image Analysis}, 79:102447, 2022.

\bibitem{54}
Kaiping Wang, Yan Wang, Bo~Zhan, Yujie Yang, Chen Zu, Xi~Wu, Jiliu Zhou, Dong
  Nie, and Luping Zhou.
\newblock An efficient semi-supervised framework with multi-task and curriculum
  learning for medical image segmentation.
\newblock {\em International journal of neural systems}, 32(09):2250043, 2022.

\bibitem{55}
Lequan Yu, Shujun Wang, Xiaomeng Li, Chi-Wing Fu, and Pheng-Ann Heng.
\newblock Uncertainty-aware self-ensembling model for semi-supervised 3d left
  atrium segmentation.
\newblock In {\em Medical Image Computing and Computer Assisted
  Intervention--MICCAI 2019: 22nd International Conference, Shenzhen, China,
  October 13--17, 2019, Proceedings, Part II 22}, pages 605--613. Springer,
  2019.

\bibitem{56}
Jizong Peng, Guillermo Estrada, Marco Pedersoli, and Christian Desrosiers.
\newblock Deep co-training for semi-supervised image segmentation.
\newblock {\em Pattern Recognition}, 107:107269, 2020.

\bibitem{57}
Lihe Yang, Wei Zhuo, Lei Qi, Yinghuan Shi, and Yang Gao.
\newblock St++: Make self-training work better for semi-supervised semantic
  segmentation.
\newblock In {\em Proceedings of the IEEE/CVF Conference on Computer Vision and
  Pattern Recognition}, pages 4268--4277, 2022.

\end{thebibliography}

\end{document}